\documentclass[11pt,a4paper]{article}
\usepackage[hyperref]{emnlp2018}
\usepackage{times}
\usepackage{latexsym}

\usepackage{url}

\usepackage{amsmath}
\usepackage{booktabs}
\usepackage{float}
\usepackage{graphicx}
\usepackage{tipa}
\usepackage{CJKutf8}
\usepackage[utf8]{inputenc}
\usepackage[vietnam, english]{babel}
\usepackage{xcolor}

\usepackage{url}

\aclfinalcopy 
\def\aclpaperid{601} 

\title{Multimodal neural pronunciation modeling for spoken languages \\with logographic origin}

\author{Minh Nguyen \\
  National University of \\
  Singapore \\
  {\tt elenguy@nus.edu.sg} \\\And
  Gia H. Ngo \\
  National University of \\
  Singapore \\
  {\tt ngohgia@u.nus.edu} \\\And
  Nancy F. Chen \\
  Institute for Infocomm Research \\
  Singapore \\
  {\tt nancychen@alum.mit.edu} \\
}

\date{}

\begin{document}
\maketitle
\begin{abstract}
Graphemes of most languages encode pronunciation, though some are more explicit than others.
Languages like Spanish have a straightforward mapping between its graphemes and phonemes,
while this mapping is more convoluted for languages like English.
Spoken languages such as Cantonese present even more challenges in pronunciation modeling:
(1) they do not have a standard written form,
(2) the closest graphemic origins are logographic Han characters, of which only a subset of these logographic characters implicitly encodes pronunciation.
In this work, we propose a multimodal approach to predict the pronunciation of Cantonese logographic characters, using neural networks with a geometric representation of logographs and pronunciation of cognates in historically related languages.
The proposed framework improves performance by 18.1\% and 25.0\% respective to unimodal and multimodal baselines. 
\end{abstract}

\section{Introduction}
In phonographic languages, there is a direct correspondence between graphemes and phonemes~\cite{defrancis1996graphemic}, though this correspondence is not always one-to-one.
For example, in English, the word \emph{table} corresponds to the pronunciation \textipa{[``teI.bl]}, in which each alphabetic character corresponds to one phoneme, and the character \textit{e} is mapped to silence.
However, in logographic languages, the correspondence between graphemes and phonemes is more ambiguous~\cite{defrancis1996graphemic},
as only some sub-units in a grapheme are indicative of its phonemes.
Korean\footnote{A large portion of Korean vocabulary are Sino-Korean written in Hanja (Korean logographs)~\cite{sohn2001korean}},
Vietnamese\footnote{Traditional Vietnamese vocabulary comprises of Sino-Vietnamese words written by Chinese logographs and locally-invented Nom logographs~\cite{alves1999s}.}
and Chinese languages (e.g. Cantonese) are examples of logographic languages,
all belonging to the Han logographic family.
Similar to pronunciation modeling in phonographic languages, in which words are broken down into characters and modeling is done at the character level,
pronunciation modeling in logographic languages requires decomposing logographs into sub-units and extracting only sub-units carrying pronunciation hints.
As the correspondence of Han logograph to phoneme is intricately complex with many sub-rules or exceptions~\cite{hashimoto1978current},
it is challenging to computationally model these correspondences using white box approaches (e.g. graphical model).
Instead, we exploit neural networks, as they
(1) can flexibly model the implicit similarity of grapheme-phoneme relationships across languages with Han origin,
(2) can automatically learn the most relevant knowledge representation with minimal feature engineering~\cite{lecun2015deep}, such as extracting pronunciation hints from logographic representations.

Due to historical contact, there is much lexical overlap across Han logographic languages, as they borrowed words from one another~\cite{rokuro1969chinese,miyake1997pre,loveday1996language,sohn2001korean,alves1999s}.
As a result, cognates in different languages are written using identical graphemes but pronounced differently.
For example, \textipa{[she]} in Mandarin and \textipa{[sip]} in Cantonese are cognates; their pronunciations are different yet they are written using the same logograph (\begin{CJK*}{UTF8}{bsmi}懾\end{CJK*}), which represents ``admire''.
Though Han logographic languages are mutually unintelligible~\cite{tang2009mutual,handel2015classification}, the correspondence of Han logographic graphemes to phonemes across languages is often similar in systematic ways~\cite{cai2011lexical,frellesvig2008japanese,miyake1997pre}.
The shared characteristics in pronunciation of cognates could be leveraged in deciphering the pronunciation of Han logographs.
In this work, we proposed a neural pronunciation model that exploits both embeddings of logographs and cognates' phonemes.
The proposed model significantly improves pronunciation prediction of logographs in Cantonese.

\section{Related Work}
The basic units in writing (graphemes) of Han logographic languages are logographs.
A word contains one or more logographs and a logograph consists of one or more radicals.
The pronunciation of a logograph corresponds to a syllable which has three phonemes: onset, nucleus and coda.

Grapheme-to-phoneme (G2P) approaches such as~\cite{xu2004grapheme,chen2016acoustic} predicted a Han logograph's pronunciation from its local context in a phrase.
This was similar to predicting a Latin word's pronunciation from its surrounding words, essentially treated individual logographs as the basic units of the model and did not delve further into the logographic sub-units (the radicals).

While we are unaware of any work that derives features for pronunciation prediction from logographs,
there are recent work in deriving representation of logographs for various semantic tasks.
Some methods ~\cite{shi2015radical,ke2017radical,nguyen2017sub,zhuang2017natural} decomposed logographs into sub-units using expert-defined rules and then extracted the relevant semantic features.
Other methods use convolutional neural network to extract features from the images of logographs~\cite{dai2017glyph, liu2017learning, toyama2017utilizing}.
Other works combined multiple level of information for feature extraction,
using both logograph and sub-units obtained from logograph decomposition~\cite{dong2016character,han2017dual,peng2017radical,yu2017joint,yin2016multi}.

In this work, we explicitly looked at the relationship between a logograph's constituent radicals and its pronunciation.
Among Han logographs, 81\% of frequently used logographs are semantic-phonetic compounds~\cite{li1993analysis} which consist of radicals that might contain phonetic or semantic hints~\cite{hsiao2006analysis}.
The pronunciation of a logograph could conceivably be predicted from the phonetic radicals.
Furthermore, the relative position of radicals in the logograph might also offer clues about it pronunciation.
Table~\ref{table_radical_position} shows an example of such intricate relationships between a logograph's pronunciation and its constituent radicals.
All Han logographs in the table have a common phonetic radical (in red), which offers an inkling of the pronunciation of these logographs. \begin{CJK*}{UTF8}{bkai}咅\end{CJK*}
For instance, logographs that have the phonetic radical on the left (\begin{CJK*}{UTF8}{bsmi}剖\end{CJK*} and \begin{CJK*}{UTF8}{bsmi}部\end{CJK*}) share a similar pronunciation in Korean (in blue) while logographs that have the phonetic radical on the right (\begin{CJK*}{UTF8}{bsmi}陪\end{CJK*}, \begin{CJK*}{UTF8}{bsmi}賠\end{CJK*}, and \begin{CJK*}{UTF8}{bsmi}蓓\end{CJK*}) share a similar pronunciation in Mandarin, Cantonese and Vietnamese.
Note that for each logograph, their pronunciations across the different languages share similarities: when the phonetic radical is on the left, the nucleus ends in a back vowel like \textit{u} or \textit{o}, whereas when the phonetic radical is on the right, the nucleus ends in a front vowel like \textit{i}. 
\begin{table}[H]
\centering
\vspace{-3mm}
\centerline{\includegraphics[width=7.9cm]{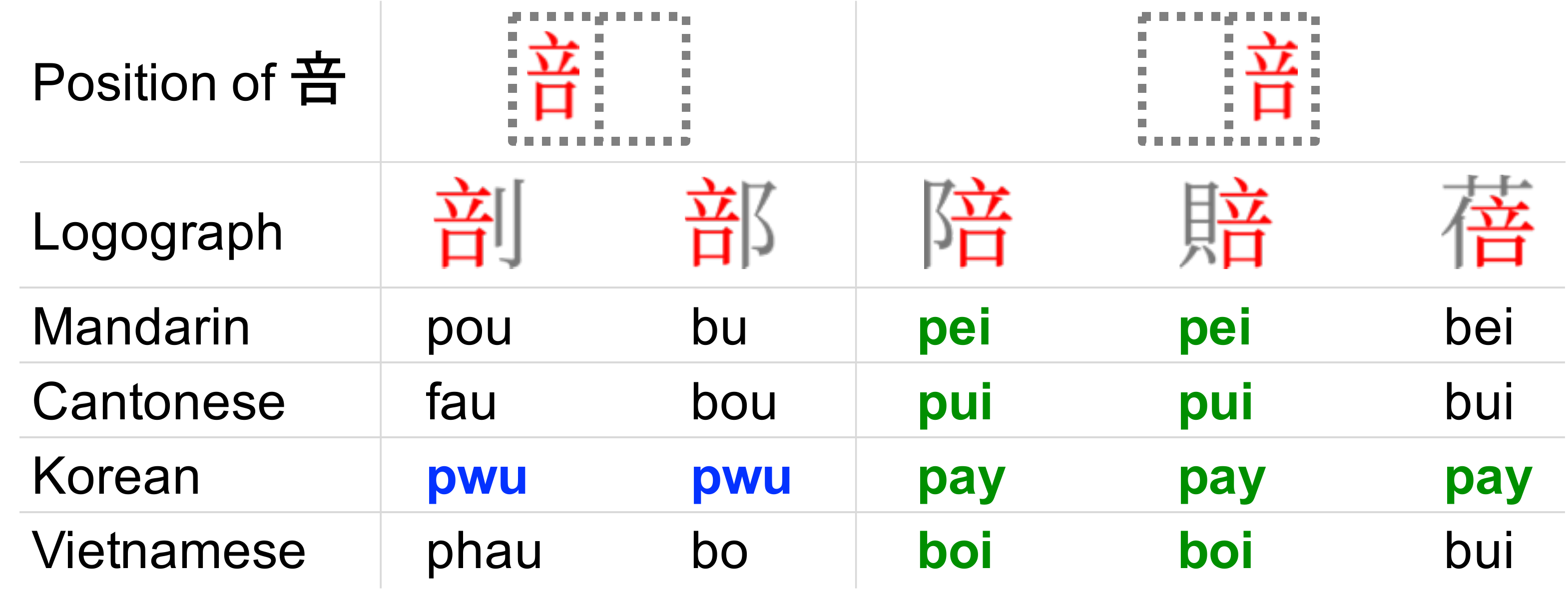}}
\vspace*{-9mm}
\textit{\caption{\label{table_radical_position}
    The position of radicals affects pronunciations. All logographs share a common radical in red. Similar pronunciations for \begin{CJK*}{UTF8}{bsmi}剖\end{CJK*} and \begin{CJK*}{UTF8}{bsmi}部\end{CJK*} are bolded in blue. Similar pronunciations for \begin{CJK*}{UTF8}{bsmi}陪\end{CJK*}, \begin{CJK*}{UTF8}{bsmi}賠\end{CJK*}, and \begin{CJK*}{UTF8}{bsmi}蓓\end{CJK*} are bolded in green. The  pronunciation of a logograph in Mandarin, Cantonese, Korean and Vietnamese are represented by
Pinyin, Jyutping, Yale, and Vietnamese alphabet symbols respectively.
}}
\vspace{-4mm}
\end{table}
The example in Table~\ref{table_radical_position} explains the motivation for our proposed approach to predict a logograph's pronunciation by modelling both the constituent radicals and their geometric positions.
Furthermore, the proposed approach can generalize to unseen logographs if the co-occurrence patterns of their constituent radicals have been learnt.

\section{Model}
We first describe a geometric decomposition of logographs and then different neural pronunciation models for logographs.
Finally, we present a multimodal neural model that incorporates both logographic input and the cognates' phonemes in predicting pronunciation of logographs.

\subsection*{Representation of Han logographs}
The majority of logographs (characters) in Han logographic language family comprise of a radical that indicates its nominal semantic category and a phonetic radical that gives an inkling of the pronunciation~\cite{defrancis1996graphemic}.
Thus, patterns of co-occurrence of radicals across logographs might be exploited to find the phonetic radicals, which in turn can suggest the corresponding pronunciation of a logograph.
Using this intuition, we model the pronunciation of logographs at the radical level.

We investigated two representations of radicals in a logograph.
In the first approach, a logograph is represented as a bag of its unordered constituent radicals (BoR), encoded as a vector of radical counts.
The second approach is to use a decomposition of radicals in the logograph that retains the original geometric organization of the radicals.
The geometric decomposition (GeoD) approach preserves important cues about the word's pronunciation in the relative position of the radicals.
For example, differentiating the left radical from the right radical in a left-right semantic-phonetic compound allows more effective extraction of pronunciation hints.
In addition, radicals that should be interpreted together are closer spatially in the GeoD representation, making the knowledge representation easier to learn.
Note that the GeoD representation is lossless as the original logograph can be reconstructed perfectly (details in Appendix~\ref{appendix_ids_algo}).
Figure~\ref{fig_logo_tree} shows the geometric decomposition of the Han logograph ``admire'' at three levels of granularity.

\begin{figure}[t]
\centering
\centerline{\includegraphics[width=7.9cm]{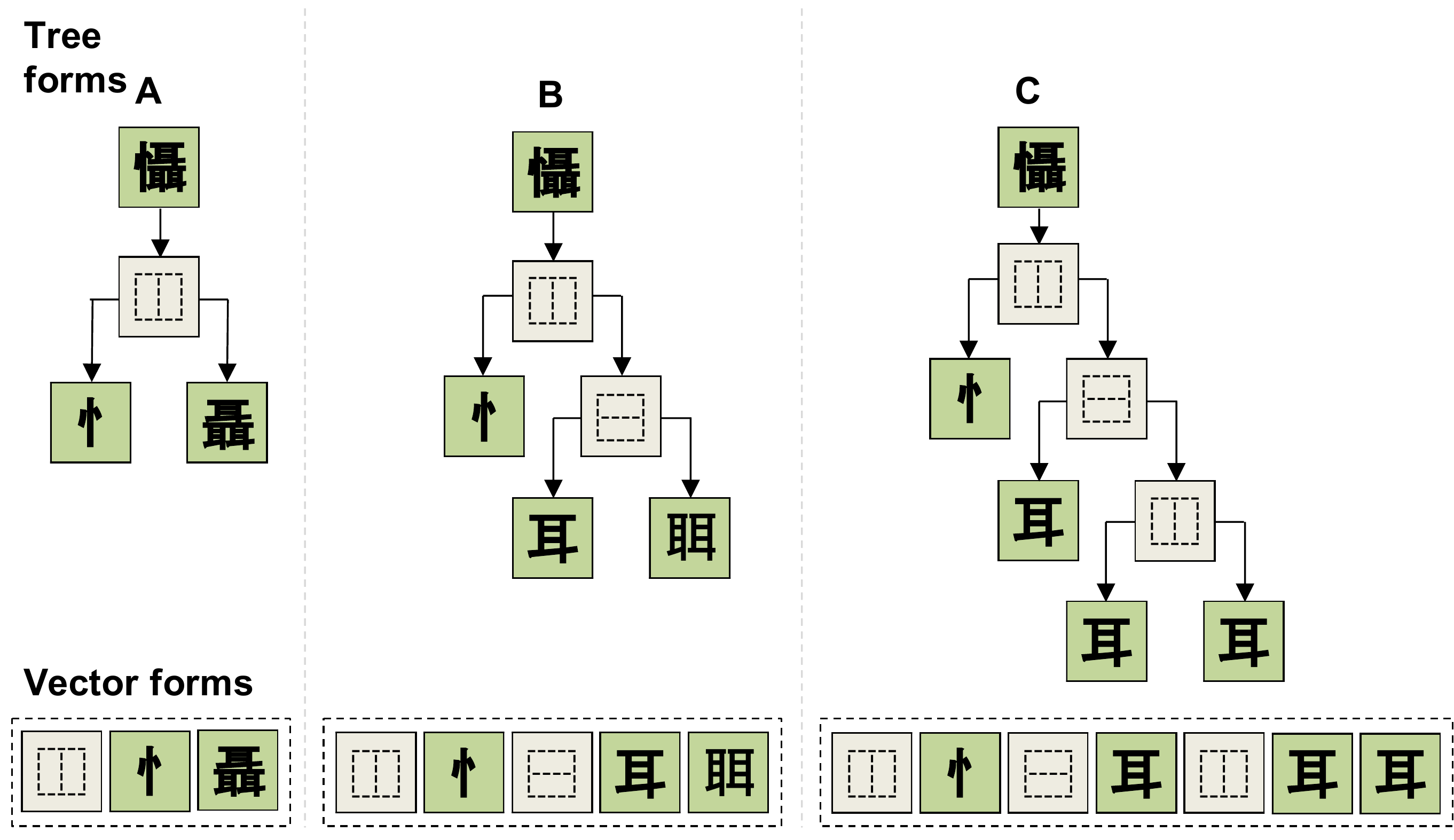}}
\vspace*{-9mm}
\textit{\caption{\label{fig_logo_tree}
    Geometric representation of the logograph ``admire''. \textbf{A}, \textbf{B} and \textbf{C} are equivalent decomposition of the same logograph but with different levels of granularity. The geometric representation comprises of both the radicals and geometric operators, which can be used to reconstruct the original logograph.
}}
\vspace{-6mm}
\end{figure}

\subsection*{Neural pronunciation prediction models}
\label{subsec:multi-modal_model}
\begin{figure}[ht]
\centering
\centerline{\includegraphics[width=8cm]{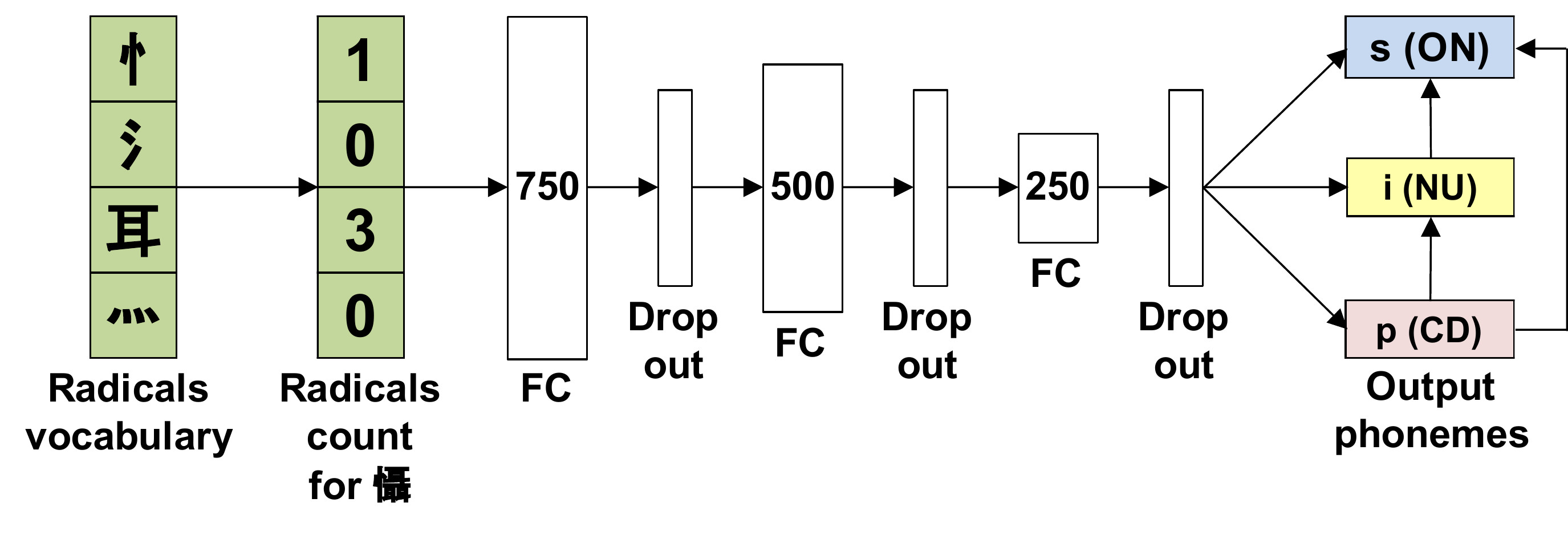}}
\vspace*{-10mm}
\textit{\caption{\label{fig_bag_of_words_model}
    Pronunciation model of logographs using multilayer perceptron (MLP). FC: Fully connected.
}}
\vspace*{-4mm}
\end{figure}

\begin{figure}[ht]
\centering
\centerline{\includegraphics[width=8cm]{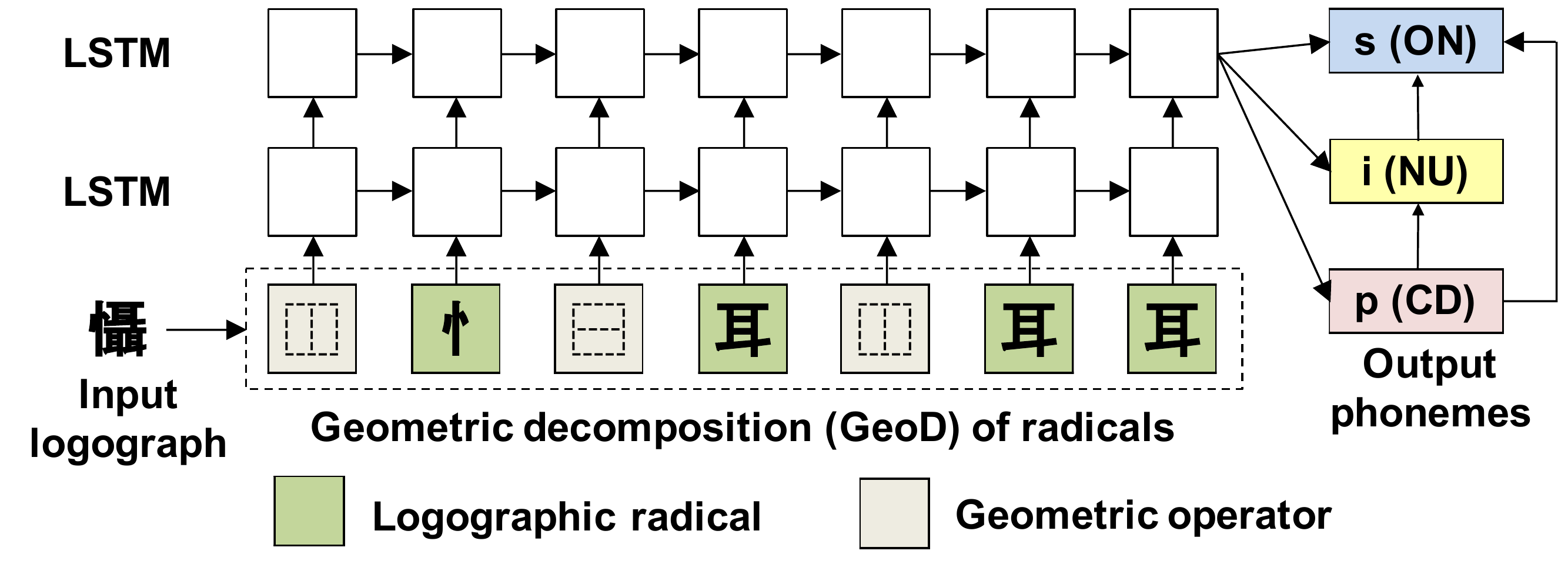}}
\vspace*{-8mm}
\textit{\caption{\label{fig_geometric_model}
    Neural pronunciation model with geometric decomposition of logographs.
}}
\vspace*{-5mm}
\end{figure}

Figure~\ref{fig_bag_of_words_model} and Figure~\ref{fig_geometric_model} show two neural pronunciation prediction models of logographs.
In Figure~\ref{fig_bag_of_words_model}, each logograph is treated as an ordered ``bag of radicals'' (BoR).
For example, assume the vocabulary of radicals in the whole dataset is [\begin{CJK*}{UTF8}{gbsn}忄, 氵, 耳, 灬\end{CJK*}], the word \begin{CJK*}{UTF8}{bsmi}懾\end{CJK*} (``admire'' - see Figure~\ref{fig_logo_tree}) is represented by a vector of counts $[1, 0, 3, 0]$, corresponding to one radical \begin{CJK*}{UTF8}{gbsn}忄\end{CJK*} and three radicals \begin{CJK*}{UTF8}{gbsn}耳\end{CJK*}.
The BoR is input to a multilayer perceptron (MLP) with three layers of size 750, 500, 250.
L2 regularization of 1e-4 is applied to the hidden layers.
The three dropout layers have dropout probabilities of 0.5, 0.5, and 0.2, respectively.
As the output variables are categorical, cross-entropy loss was used.

We investigated two structures for predicting output phonemes (i.e.\ onset, nucleus, coda).
In the first structure, output phonemes were predicted independently using the last hidden layer.
The second structure made a sequential prediction (1) the coda was first predicted using the last hidden layer (2) the nucleus was predicted using both the final hidden layer and the predicted coda, and (3) the onset was predicted using the last hidden layer together with the predicted coda and nucleus.
The second structure was motivated by a stronger dependency between the nuclues and coda. For example, the nucleus and coda are often grouped together as a single unit (rime/final) in the syllabic structure of most languages \cite{kessler2002syllable}.
In our experiments, the sequential structure yielded lower error rates so it is used in all neural network models.

In Figure~\ref{fig_geometric_model}, each logograph is represented by its geometric decomposition (GeoD).
For example, the logograph \begin{CJK*}{UTF8}{bsmi}懾\end{CJK*} is represented by a sequence of radicals and geometric operators shown in Figure~\ref{fig_logo_tree}C.
The neural prediction model consists of two LSTM layers with 256 memory cells each.
Input and recurrent dropout~\cite{gal2016theoretically} of 0.2 and 0.5 are applied to the LSTM layers to prevent overfitting.

\subsection*{Multimodal neural pronunciation model of logographs}
\begin{figure*}[ht]
\vspace*{-7mm}
\centering
\centerline{\includegraphics[width=16cm]{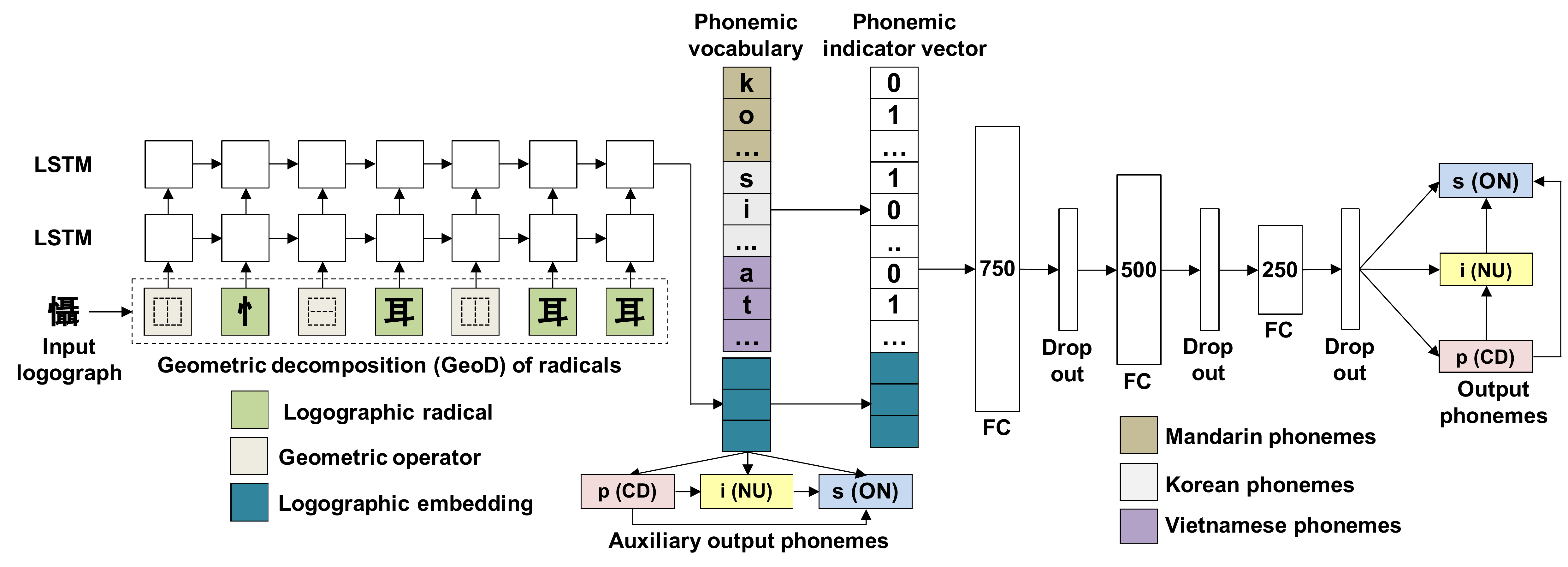}}
\vspace*{-8mm}
\textit{\caption{\label{fig_multimodal_model}
Multimodal neural pronunciation prediction model using logographs' geometric representation and cognates' phonemes.
}}
\vspace*{-4mm}
\end{figure*}

In this section, we want to model the pronunciations of a logograph in the target language Cantonese using multimodal information from both the logograph and phonemes of the cognates, as shown in Figure~\ref{fig_multimodal_model}.
Given a vocabulary of phonemes in the source languages related to Cantonese (Mandarin, Korean, Vietnamese), the cognates' phonemes are encoded as an indicator vector, with an element equals 1 if the corresponding phoneme in the vocabulary appears in a cognate's pronunciation, and 0 otherwise.

The geometric decomposition (GeoD) of the logograph is fed to two LSTM layers.
The output at the last time step is concatenated together with the multilingual phonemic vector and used as input for a multi-layer perceptron (MLP).
The MLP and LSTM setups are the same as those in Figure~\ref{fig_bag_of_words_model} and Figure~\ref{fig_geometric_model} respectively.
Deep supervision~\cite{Szegedy_2015_CVPR} was applied by using the output of the LSTM to make auxiliary prediction of the output phonemes.
Note that the auxiliary prediction should be identical to the main prediction.
While predicting the same target, the main prediction used both cognate phonemes and the logograph while the auxiliary prediction used only the logograph.
This was to ensure features extracted from the logographs are useful for pronunciation prediction and are complementary to the features extracted from the multilingual phonemes.

\label{subsect_multi_modal_model}

\section{Experiments}
We investigate whether Cantonese phonemes could be predicted using Han logographs and the cognates' phonemes from Mandarin, Korean, and Vietnamese.
The prediction output are Cantonese onsets, nuclei and codas.
The experimental design is motivated by the nature of Han-logographic languages.
A Chinese logograph (character) is phonologically equivalent to a syllable in English while the constituent radicals are analogous to alphabet letters (with far less phonetic information).
While in most languages, a syllable’s pronunciation is influenced by neighboring syllables, most Han-logographic languages are monosyllabic and a logograph’s pronunciation is rarely affected by neighboring logographs.
Therefore, pronunciation prediction at the logograph (character) level for Han logographs is more appropriate.
We use string error rate (SER) and token error rate (TER) as evaluation metrics.
A wrongly predicted phoneme (onset, nucleus or coda) is counted as one token error.
A syllable containing token error(s) is counted as one string error.
All the neural networks were trained using Adam~\cite{kingma_adam:_2014}.

\subsection*{Data}
The dataset is extracted from the UniHan database,\footnote{https://www.unicode.org/charts/unihan.html}
which is a pronunciation database of logographs from Han logographic languages and maintained by the Unicode consortium.
For each entry in the dataset, a logograph corresponds to phonemes in Cantonese, Mandarin, Korean and Vietnamese, represented by
Jyutping,\footnote{https://en.wikipedia.org/wiki/Jyutping}
Pinyin,\footnote{https://en.wikipedia.org/wiki/Pinyin}
Yale,\footnote{https://en.wikipedia.org/wiki/Yale\_romanization\_of\_Korean}
and Vietnamese alphabet symbols respectively.\footnote{https://en.wikipedia.org/wiki/Vietnamese\_alphabet}
We randomly partition the dataset into two sets, with 80\% for training and the other 20\% for testing. 
Overall, there are 16,011 entries in the training set and 4,002 entries in the test set.
1000 entries of the training set are used as the development set for hyper-parameters fine-tuning.

In the test set, only 16\% of logographs have pronunciations in all non-target languages, while 6\% of logographs have no non-target language pronunciation.
The availability of pronunciations in non-target languages differs from logograph to logograph.
For example, some logographs have Mandarin and Korean pronunciations, while others only have Mandarin pronunciations.

\subsection*{Predicting pronunciation using logograph input}

We compared the neural networks against a decision tree baseline.
The decision tree baseline was implemented using scikit-learn \cite{scikit-learn}.
The input of the decision tree (DT) model is the BoR representation of the logograph, while the input of neural networks can be either BoR or GeoD.
The MLP network in Figure~\ref{fig_bag_of_words_model} uses BoR, while the LSTM in Figure~\ref{fig_geometric_model} uses GeoD as input.
All models output phonemes in Cantonese.

From Table~\ref{table_ids}, the neural network (MLP) outperforms decision tree when using BoR input.
Both the SER and TER of the MLP model are lower than those of the decision tree.
The LSTM model using GeoD leads to the lowest SER and TER,
suggesting the benefits of relative positional information of radicals in predicting pronunciation.
The trends of onset, nucleus and coda error rates are similar to those of TER and SER.
However, as the gap of of error rate between MLP (BoR) and LSTM (GeoD) for TER and SER are quite small, using BoR instead of GeoD can be a good computation-accuracy trade-off.

\begin{table}[H]
\centering
\vspace{-2mm}
\begin{tabular}{l@{\hspace{1em}}r@{\hspace{0.5em}}r@{\hspace{1em}}r@{\hspace{0.5em}}r@{\hspace{0.5em}}r}
\toprule {\bfseries}
    Method & SER & TER & On. & Nu. & Cd. \\
\midrule
    DT (BoR) & 63.8 & 39.8 & 50.7 & 45.7 & 22.9 \\
\midrule
    MLP (BoR) & 59.2 & 33.6 & 44.5 & 38.6 & 17.8 \\
\midrule
    LSTM (GeoD) & \textbf{58.4} & \textbf{32.6} & \textbf{43.3} & \textbf{37.4} & \textbf{17.1} \\
\bottomrule
\end{tabular}
\textit{\caption{\label{table_ids}
    Prediction error rates of Cantonese phonemes by decision tree (DT),
    MLP and LSTM using only logographic input.
    Best results are in bold.
}}
\end{table}
\vspace{-6mm}

\subsection*{Predicting pronunciation using multimodal input}
The input of the models are logographs and cognate phonemes from Mandarin, Korean and Vietnamese.
Table~\ref{table_multimodal} shows that the proposed multimodal neural network exploits multimodal and geometric information effectively.
The relative improvement reaches 18.2\% and 33.3\% for SER and TER respectively. 
The last rows in Table~\ref{table_ids} and Table~\ref{table_multimodal} show that by combining Korean, Mandarin and Vietnamese phonemes input with GeoD, the prediction performance improves by 54.1\% relative in TER and by 65.5\% relative in SER.
Moreover, using solely logograph input resulted in higher onset error (43.3\%) than nucleus error (37.4\%) while using both logographs and multilingual phonemes improves the onset error (23.5\%) to be lower than nucleus error (24.6\%).
This suggests that logographs and phonemes of cognates provide complementary information about the pronunciation of a logograph, which in this case, most notably at the onset position.
While logographs usually carry hints about phonemes at the nucleus and coda position but not at the onset position, multilingual phonemes input might carry hints about pronunciation at all three positions.
\begin{table}[H]
\centering
\vspace{-2mm}
\begin{tabular}{l@{\hspace{1em}}r@{\hspace{0.5em}}r@{\hspace{1em}}r@{\hspace{0.5em}}r@{\hspace{0.5em}}r}
\toprule {\bfseries}
    Method & SER & TER & On. & Nu. & Cd. \\
\midrule
    DT (BoR, ph) & 44.0 & 24.8 & 29.8 & 29.9 & 14.7 \\
\midrule
    MLP (BoR, ph) & 38.5 & 19.6 & 23.4 & 24.8 & 10.5 \\
\midrule
    LSTM (GeoD, ph) & \textbf{37.2} & \textbf{18.6} & \textbf{22.6} & \textbf{23.4} & \textbf{9.8} \\
\bottomrule
\end{tabular}
\vspace{-2mm}
\textit{\caption{\label{table_multimodal}
    Prediction error rates of Cantonese phonemes by multimodal models;
    BoR: Bag of Radicals; GeoD: Geometric Decomposition; ph:phonemes.
    Best results are in bold.
}}
\end{table}
\vspace{-6mm}

\section{Discussion}
We have empirically shown that the systematic yet tenuous correspondence between pronunciations of cognates in Han logographic languages can be exploited for pronunciation modeling using neural networks. 
Moreover, combining logograph with cognate pronunciations further improves pronunciation prediction. These results could be potentially applied to speech processing tasks such as speech synthesis, where the construction of pronunciation dictionaries are expert labor-intensive, especially for under-resourced spoken languages.

For future work, recursive neural network~\cite{tai2015improved} can be used as it is better suited for the hierarchical logographic decomposition.
Besides, incorporating more detailed relationship between radicals (e.g.~\cite{zhuang2017natural}) can help improve the model. 
The proposed approaches can also be applied to other languages such as Min Nan or Hakka, which are spoken languages that are even less well-documented than Cantonese. 

\bibliography{emnlp2018}

\begin{thebibliography}{34}
\expandafter\ifx\csname natexlab\endcsname\relax\def\natexlab#1{#1}\fi

\bibitem[{Alves(1999)}]{alves1999s}
Mark~J Alves. 1999.
\newblock {What’s so Chinese about Vietnamese}.
\newblock In \emph{Papers from the ninth annual meeting of the Southeast Asian
  Linguistics Society}, pages 221--242.

\bibitem[{Cai et~al.(2011)Cai, Pickering, Yan, and Branigan}]{cai2011lexical}
Zhenguang~G Cai, Martin~J Pickering, Hao Yan, and Holly~P Branigan. 2011.
\newblock {Lexical and syntactic representations in closely related languages:
  Evidence from Cantonese--Mandarin bilinguals}.
\newblock \emph{Journal of Memory and Language}, 65(4):431--445.

\bibitem[{Chen et~al.(2016)Chen, Povey, and Khudanpur}]{chen2016acoustic}
Guoguo Chen, Daniel Povey, and Sanjeev Khudanpur. 2016.
\newblock Acoustic data-driven pronunciation lexicon generation for logographic
  languages.
\newblock In \emph{Acoustics, Speech and Signal Processing (ICASSP), 2016 IEEE
  International Conference on}, pages 5350--5354. IEEE.

\bibitem[{Dai and Cai(2017)}]{dai2017glyph}
Falcon~Z Dai and Zheng Cai. 2017.
\newblock {Glyph-aware Embedding of Chinese Characters}.
\newblock \emph{EMNLP 2017}, page~64.

\bibitem[{Defrancis(1996)}]{defrancis1996graphemic}
John Defrancis. 1996.
\newblock Graphemic indeterminacy in writing systems.
\newblock \emph{Word}, 47(3):365--377.

\bibitem[{Dong et~al.(2016)Dong, Zhang, Zong, Hattori, and
  Di}]{dong2016character}
Chuanhai Dong, Jiajun Zhang, Chengqing Zong, Masanori Hattori, and Hui Di.
  2016.
\newblock {Character-based LSTM-CRF with radical-level features for Chinese
  named entity recognition}.
\newblock In \emph{Natural Language Understanding and Intelligent
  Applications}, pages 239--250. Springer.

\bibitem[{Frellesvig and Whitman(2008)}]{frellesvig2008japanese}
Bjarke Frellesvig and John Whitman. 2008.
\newblock {The Japanese-Korean vowel correspondences}.
\newblock \emph{Japanese/Korean Linguistics}, 13:15--28.

\bibitem[{Gal and Ghahramani(2016)}]{gal2016theoretically}
Yarin Gal and Zoubin Ghahramani. 2016.
\newblock A theoretically grounded application of dropout in recurrent neural
  networks.
\newblock In \emph{Advances in neural information processing systems}, pages
  1019--1027.

\bibitem[{Han et~al.(2017)Han, Xiaokun, Lei, Hua, Zhimin, Yi, and
  George}]{han2017dual}
He~Han, Yang Xiaokun, Wu~Lei, Yan Hua, Gao Zhimin, Feng Yi, and Townsend
  George. 2017.
\newblock Dual long short-term memory networks for sub-character representation
  learning.
\newblock \emph{arXiv preprint arXiv:1712.08841}.

\bibitem[{Handel(2015)}]{handel2015classification}
Zev Handel. 2015.
\newblock {The classification of Chinese: sinitic (the Chinese language
  family)}.
\newblock In \emph{The Oxford handbook of Chinese linguistics}, pages 34--44.
  Oxford University Press.

\bibitem[{Hashimoto(1978)}]{hashimoto1978current}
Mantaro~J Hashimoto. 1978.
\newblock {Current developments in Sino—Vietnamese studies}.
\newblock \emph{Journal of Chinese Linguistics}, pages 1--26.

\bibitem[{Hsiao and Shillcock(2006)}]{hsiao2006analysis}
Janet Hui-wen Hsiao and Richard Shillcock. 2006.
\newblock Analysis of a chinese phonetic compound database: Implications for
  orthographic processing.
\newblock \emph{Journal of psycholinguistic research}, 35(5):405--426.

\bibitem[{Ke and Hagiwara(2017)}]{ke2017radical}
Yuanzhi Ke and Masafumi Hagiwara. 2017.
\newblock {Radical-level Ideograph Encoder for RNN-based Sentiment Analysis of
  Chinese and Japanese}.
\newblock \emph{arXiv preprint arXiv:1708.03312}.

\bibitem[{Kessler and Treiman(2002)}]{kessler2002syllable}
Brett Kessler and Rebecca Treiman. 2002.
\newblock Syllable structure and the distribution of phonemes in english
  syllables.

\bibitem[{Kingma and Ba(2014)}]{kingma_adam:_2014}
Diederik~P. Kingma and Jimmy Ba. 2014.
\newblock Adam: {A} {Method} for {Stochastic} {Optimization}.
\newblock \emph{arXiv:1412.6980 [cs]}.
\newblock ArXiv: 1412.6980.

\bibitem[{LeCun et~al.(2015)LeCun, Bengio, and Hinton}]{lecun2015deep}
Yann LeCun, Yoshua Bengio, and Geoffrey Hinton. 2015.
\newblock Deep learning.
\newblock \emph{Nature}, 521(7553):436.

\bibitem[{Li and Kang(1993)}]{li1993analysis}
Y~Li and JS~Kang. 1993.
\newblock {Analysis of phonetics of the ideophonetic characters in Modern
  Chinese}.
\newblock \emph{Information analysis of usage of characters in modern Chinese},
  pages 84--98.

\bibitem[{Liu et~al.(2017)Liu, Lu, Lo, and Neubig}]{liu2017learning}
Frederick Liu, Han Lu, Chieh Lo, and Graham Neubig. 2017.
\newblock Learning character-level compositionality with visual features.
\newblock \emph{arXiv preprint arXiv:1704.04859}.

\bibitem[{Loveday(1996)}]{loveday1996language}
Leo~J Loveday. 1996.
\newblock \emph{Language contact in Japan: A sociolinguistic history}.
\newblock Clarendon Press.

\bibitem[{Miyake(1997)}]{miyake1997pre}
Marc~Hideo Miyake. 1997.
\newblock {Pre-Sino-Korean and Pre-Sino-Japanese: reexamining an old Problem
  from a modern perspective}.
\newblock \emph{Japanese/Korean Linguistics}, 6:179--211.

\bibitem[{Nguyen et~al.(2017)Nguyen, Brooke, and Baldwin}]{nguyen2017sub}
Viet Nguyen, Julian Brooke, and Timothy Baldwin. 2017.
\newblock {Sub-character Neural Language Modelling in Japanese}.
\newblock In \emph{Proceedings of the First Workshop on Subword and Character
  Level Models in NLP}, pages 148--153.

\bibitem[{Pedregosa et~al.(2011)Pedregosa, Varoquaux, Gramfort, Michel,
  Thirion, Grisel, Blondel, Prettenhofer, Weiss, Dubourg, Vanderplas, Passos,
  Cournapeau, Brucher, Perrot, and Duchesnay}]{scikit-learn}
F.~Pedregosa, G.~Varoquaux, A.~Gramfort, V.~Michel, B.~Thirion, O.~Grisel,
  M.~Blondel, P.~Prettenhofer, R.~Weiss, V.~Dubourg, J.~Vanderplas, A.~Passos,
  D.~Cournapeau, M.~Brucher, M.~Perrot, and E.~Duchesnay. 2011.
\newblock Scikit-learn: Machine learning in {P}ython.
\newblock \emph{Journal of Machine Learning Research}, 12:2825--2830.

\bibitem[{Peng et~al.(2017)Peng, Cambria, and Zou}]{peng2017radical}
Haiyun Peng, Erik Cambria, and Xiaomei Zou. 2017.
\newblock {Radical-based hierarchical embeddings for Chinese sentiment analysis
  at sentence level}.
\newblock In \emph{The 30th International FLAIRS conference. Marco Island}.

\bibitem[{Rokuro(1969)}]{rokuro1969chinese}
Kono Rokuro. 1969.
\newblock {The Chinese writing and its influence on the Scripts of the
  Neighbouring Peoples with special reference to Korea and Japan}.
\newblock \emph{Memoirs of the Research Department of the Toyo Bunko (The
  Oriental Library) No}, 27:117--123.

\bibitem[{Shi et~al.(2015)Shi, Zhai, Yang, Xie, and Liu}]{shi2015radical}
Xinlei Shi, Junjie Zhai, Xudong Yang, Zehua Xie, and Chao Liu. 2015.
\newblock {Radical embedding: Delving deeper to Chinese radicals}.
\newblock In \emph{Proceedings of the 53rd Annual Meeting of the Association
  for Computational Linguistics and the 7th International Joint Conference on
  Natural Language Processing (Volume 2: Short Papers)}, volume~2, pages
  594--598.

\bibitem[{Sohn(2001)}]{sohn2001korean}
Ho-Min Sohn. 2001.
\newblock \emph{The Korean Language}.
\newblock Cambridge University Press.

\bibitem[{Szegedy et~al.(2015)Szegedy, Liu, Jia, Sermanet, Reed, Anguelov,
  Erhan, Vanhoucke, and Rabinovich}]{Szegedy_2015_CVPR}
Christian Szegedy, Wei Liu, Yangqing Jia, Pierre Sermanet, Scott Reed, Dragomir
  Anguelov, Dumitru Erhan, Vincent Vanhoucke, and Andrew Rabinovich. 2015.
\newblock Going deeper with convolutions.
\newblock In \emph{The IEEE Conference on Computer Vision and Pattern
  Recognition (CVPR)}.

\bibitem[{Tai et~al.(2015)Tai, Socher, and Manning}]{tai2015improved}
Kai~Sheng Tai, Richard Socher, and Christopher~D Manning. 2015.
\newblock Improved semantic representations from tree-structured long
  short-term memory networks.
\newblock \emph{Proceedings of the 53rd Annual Meeting of the Association for
  Computational Linguistics and the 7th International Joint Conference on
  Natural Language Processing (Volume 1: Long Papers)}.

\bibitem[{Tang and Van~Heuven(2009)}]{tang2009mutual}
Chaoju Tang and Vincent~J Van~Heuven. 2009.
\newblock {Mutual intelligibility of Chinese dialects experimentally tested}.
\newblock \emph{Lingua}, 119(5):709--732.

\bibitem[{Toyama et~al.(2017)Toyama, Miwa, and Sasaki}]{toyama2017utilizing}
Yota Toyama, Makoto Miwa, and Yutaka Sasaki. 2017.
\newblock {Utilizing Visual Forms of Japanese Characters for Neural Review
  Classification}.
\newblock In \emph{Proceedings of the Eighth International Joint Conference on
  Natural Language Processing (Volume 2: Short Papers)}, volume~2, pages
  378--382.

\bibitem[{Xu et~al.(2004)Xu, Fu, and Li}]{xu2004grapheme}
Jun Xu, Guohong Fu, and Haizhou Li. 2004.
\newblock Grapheme-to-phoneme conversion for chinese text-to-speech.
\newblock In \emph{Eighth International Conference on Spoken Language
  Processing}.

\bibitem[{Yin et~al.(2016)Yin, Wang, Li, Li, and Wang}]{yin2016multi}
Rongchao Yin, Quan Wang, Peng Li, Rui Li, and Bin Wang. 2016.
\newblock {Multi-granularity Chinese word embedding}.
\newblock In \emph{Proceedings of the Conference on Empirical Methods in
  Natural Language Processing}, pages 981--986.

\bibitem[{Yu et~al.(2017)Yu, Jian, Xin, and Song}]{yu2017joint}
Jinxing Yu, Xun Jian, Hao Xin, and Yangqiu Song. 2017.
\newblock {Joint Embeddings of Chinese Words, Characters, and Fine-grained
  Subcharacter Components}.
\newblock In \emph{Proceedings of the Conference on Empirical Methods in
  Natural Language Processing}, pages 286--291.

\bibitem[{Zhuang et~al.(2017)Zhuang, Wang, Li, Wang, and
  Zhou}]{zhuang2017natural}
Hang Zhuang, Chao Wang, Changlong Li, Qingfeng Wang, and Xuehai Zhou. 2017.
\newblock {Natural Language Processing Service Based on Stroke-Level
  Convolutional Networks for Chinese Text Classification}.
\newblock In \emph{Web Services (ICWS), 2017 IEEE International Conference on},
  pages 404--411. IEEE.

\end{thebibliography}
\bibliographystyle{acl_natbib_nourl}
\clearpage

\appendix
\section{The Ideographic Description algorithm}
\label{appendix_ids_algo}
The Ideographic Description algorithm, defined by the Unicode Consortium, describes a way to represent a grapheme by its components.
All Han logographs ({\em i.e.} graphemes) can be recursively decomposed into smaller components that are themselves logographs.
With IDS denoting an logograph, the Ideographic Description algorithm can be written as
\begin{quote}
\begin{verbatim}
IDS := IDS
 | BinaryOperator IDS IDS
 | TrinaryOperator IDS IDS IDS/
\end{verbatim}
\end{quote}
This simply means that an logograph can be decomposed into one, two or three smaller logographs.
The operators indicate the relative positions of the operands.
Many logographs can be described in more than one way using this algorithm as the logographs can themselves be broken down further.

\begin{figure}[H]
\centering
\centerline{\includegraphics[width=8.5cm]{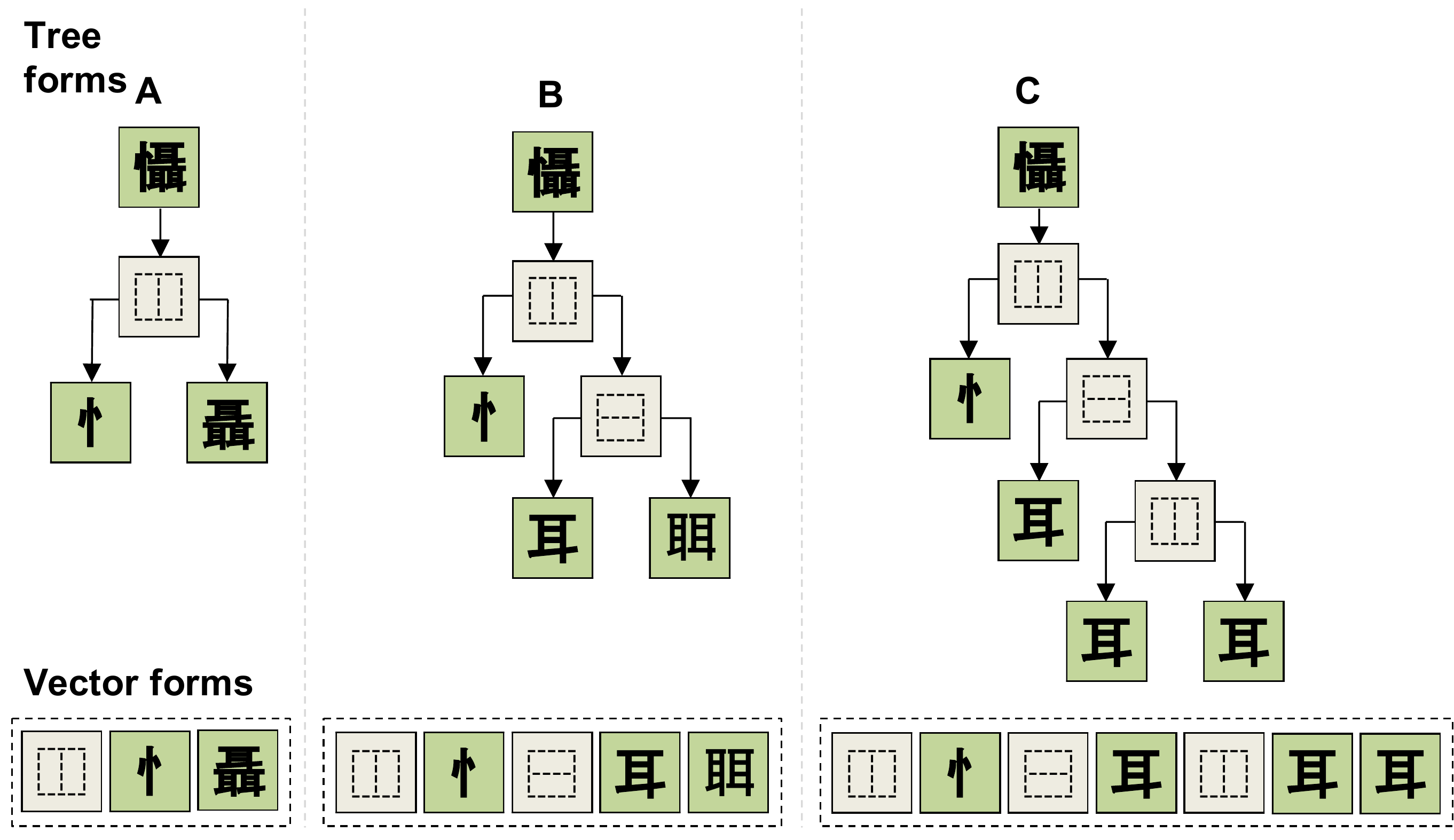}}
\makeatletter 
\renewcommand{\thefigure}{S\@arabic\c@figure}
\makeatother
\textit{\caption{\label{fig_logo_tree_appendix}
    (reproduced from Figure~\ref{fig_logo_tree})
    \textbf{A}, \textbf{B} and \textbf{C} are equivalent ideographic description sequences for the same logograph.
    Each sequence can also be represented as a tree.
}}
\end{figure}

Figure~\ref{fig_logo_tree_appendix} shows three different ways the logograph for ``admire'' can be decomposed at different levels of granularity.
The granularity depends on the set of basic logographs at which the algorithm terminates.
As the algorithm is recursive, the decomposition of a logograph is a tree or a sequence with the operators evaluated in prefix order.
The sequence representation is lossless as it preserves the relative geometric position between the components.
The logograph can be reconstructed perfectly from the sequence of components.

Figure~\ref{fig_logo_tree_appendix} also shows how the three sequence of components are represented as three vectors of count.
Different from the sequence representation, representing the components as a vector of counts is lossy, as the geometric relationship between components are not preserved.
Representing graphemes as sequences rather than as vectors may lead to higher prediction accuracy if the positional information is useful for the task.

\end{document}